\documentclass{article}
\usepackage{spconf,amsmath,graphicx}

\usepackage{amsfonts}
\usepackage{xcolor}
\usepackage{subfigure}
\usepackage[T1]{fontenc}
\usepackage{csquotes}
\usepackage{float}
\usepackage{bm}

\name{Yan Chen$^*$, James H. Holmes$^*$, Curtis Corum$^\ddag$, Vincent Magnotta$^*$, Mathews Jacob$^*$. \thanks{This work is supported by NIH R01AG067078 and 3R43MH122028-02S1, and was conducted on an instrument funded by 1S10OD025025-01.}}
\address{$^*$University of Iowa. $^\ddag$Champaign Imaging, LLC}
\begin{document}

\title{Deep Factor Model: A Novel Approach for Motion Compensated Multi-Dimensional MRI}

\maketitle

\begin{abstract}
Recent quantitative parameter mapping methods including MR fingerprinting (MRF) collect a time series of images that capture the evolution of magnetization. The focus of this work is to introduce a novel approach termed as Deep Factor Model(DFM), which offers an efficient representation of the multi-contrast image time series. The higher efficiency of the representation enables the acquisition of the images in a highly undersampled fashion, which translates to reduced scan time in 3D high-resolution multi-contrast applications. The approach integrates motion estimation and compensation, making the approach robust to subject motion during the scan.
\end{abstract}

\vspace{-1em}
\section{Introduction}
\begin{figure*}[t]
\centering
\subfigure[Low-rank representation]{
\includegraphics[width=0.25\textwidth]{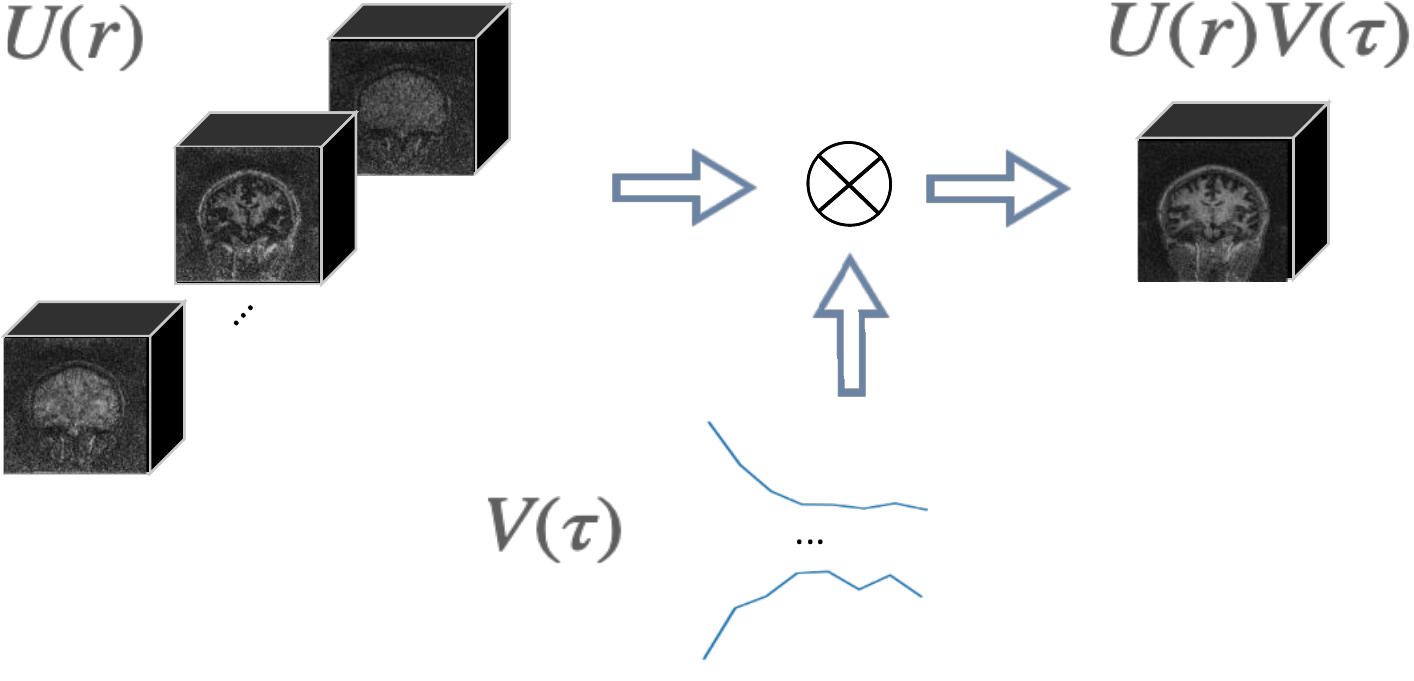}}\hspace{3em}
~~\subfigure[Proposed Deep Factor Model]{\includegraphics[width=0.5\textwidth]{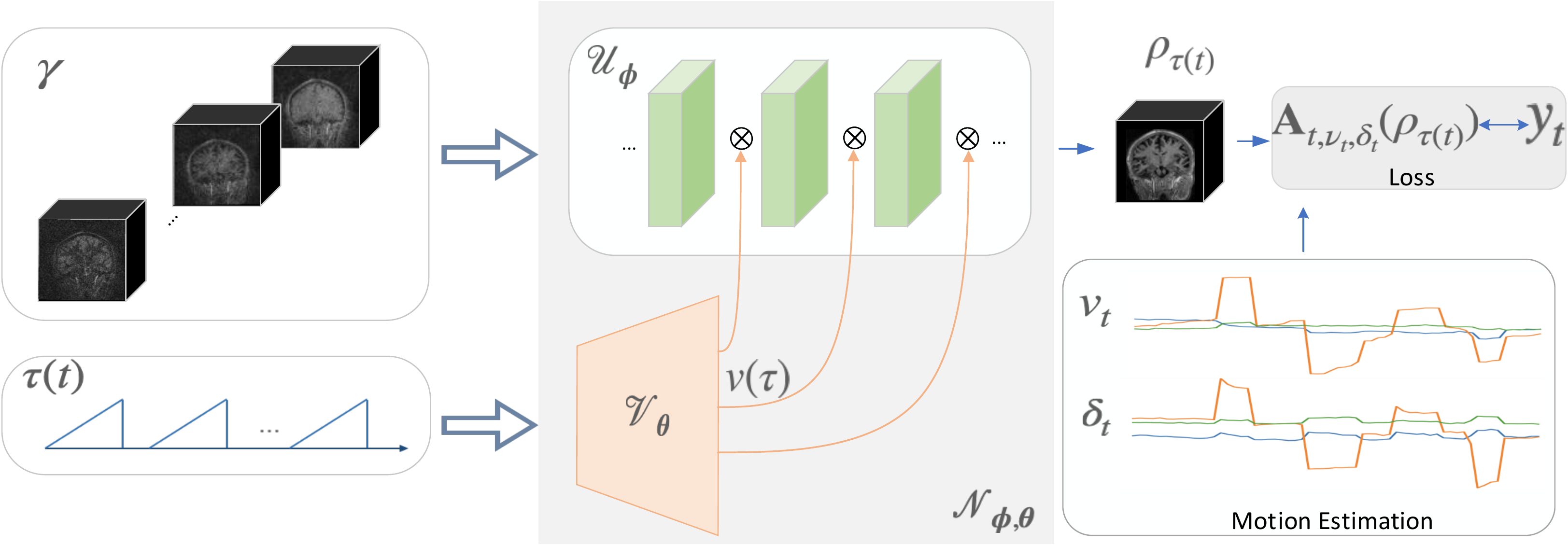}}\vspace{-1em}
\caption{Illustration of the low-rank model and DFM. (a) The low-rank method represents the image corresponding to the delay $\tau$ as the weighted linear combination of the spatial basis functions 
$\mathbf U(r)$ based on weights $\mathbf V(\tau)$; (b) DFM models the signal $\rho(\mathbf r,\tau)$ as the output of a conditional network $\mathcal N_{\phi,\theta}$. The features of the CNN $\mathcal U_{\phi}$  are modulated by $v(\tau)$ generated by a dense network $\mathcal V_{\theta}$. The input to $\mathcal V_{\theta}$ is $\tau$, while that of the $\mathcal U_{\phi}$ is an approximate initialization $\boldsymbol{\gamma}$ obtained from gridding.}\vspace{-1em}
\label{fig:dfm}
\end{figure*} 
The signal in MRI is sensitive to multiple physical properties of the tissue, which is exploited to observe organs with different contrast mechanisms. The classical MRI methods approach is to carefully choose the acquisition parameters to make the images sensitive to a specific physical property while being less sensitive to others. For example, the magnetization prepared rapid gradient echo (MPRAGE)\cite{brant1992mp} sequence uses inversion pulses and restricts the acquisition of the data at specific inversion times to obtain good contrast between the gray matter (GM) and white matter (WM) regions. However, a challenge with this approach is its low acquisition efficiency, resulting from the long waiting times and delay times to allow the magnetization to recover. The MPnRAGE\cite{kecskemeti2016mpnrage} sequence was introduced recently, which relies on continuously acquiring the images using radial acquisitions after the inversion pulse; the radial data is binned depending on the inversion time and recovered using gridding. In addition to improving acquisition efficiency, this approach enables the acquisition of images at multiple inversion times.  

The MPnRAGE approach has conceptual similarities to MRF \cite{ma2013magnetic} that continuously acquires the data using incoherent sampling patterns, while magnetization is continuously evolving. MRF attempts to estimate the physical parameters (e.g. $T_1, T_2$) directly from the undersampled k-t space data using pattern matching. These methods rely on large dictionaries of MRF derived using Bloch equation simulations, which correspond to different values of the physical parameters. Several constrained reconstruction algorithms were introduced to improve the reconstruction of the individual  frames in MRF; the strategies include low-rank factor modeling of the reconstructed images\cite{zhao2018improved}. 
% While the temporal profiles in low-rank methods can be learned from dictionaries of magnetization evolution, this approach may cause errors resulting from effects such as $B_1$ and $B_0$ inhomogeneity that are not modeled.
% A common challenge with all of the above methods is the sensitivity of the signal to subject motion during the scan, which is a key problem in non-compliant subjects, including pediatric and older adults. 

The main focus of this paper is to introduce DFM for the joint recovery of inversion recovery data. The proposed DFM, illustrated in Fig. \ref{fig:dfm} (b), is a generalization of the low-rank model. It capitalizes on the benefits of convolutional neural networks (CNN) in representing images. While unrolled algorithms \cite{sandino2021deep} were introduced to improve low-rank or subspace models, these methods need multiple copies of the subspace images at each unrolling step; the extremely high memory demand of these approaches makes them infeasible in our 3D setting. 

The DFM represents each image in the time series as the output of a conditional CNN. The input of the CNN is a series of representative image volumes corresponding to a few inversion delays obtained from gridding.
% These images are coarse approximations and often exhibit significant aliasing and blurring, resulting from undersampling.
The inversion time-dependent network derives each image in the time series as a denoised non-linear function of the gridded images. 
The temporal factors derived from a dense network are used to modulate the features of the CNN.
% We use the inversion time as the condition vector, which are used to derive the temporal factors that modulate each of the features in the CNN. The temporal factors are derived from the inversion time using a fully-connected network, inspired by adaptive instance normalization \cite{huang2017arbitrary}. 
% The parameters of both networks are jointly learned from the k-t space data. 
Unlike supervised deep learning models, DFM is an  unsupervised learning approach that directly learns the subject-specific representation from the measured k-t space data. This approach of learning the network parameters from the undersampled measurements is motivated by \cite{zou2021dynamic}, which learns an image or an ensemble of images from undersampled measurements. 
We hypothesize that non-linear deep representation is more efficient that the linear low-rank factor models, and offers implicit spatial regularization due to the implicit bias of CNN blocks towards images. 

We also capitalize on the unsupervised learning strategy to compensate for subject motion. In particular, we model the subject motion during the scan as a rigid body. We use the relation between rigid body motions and corresponding transformations in the Fourier domain to absorb them into the forward model. These time-varying  parameters are assumed to be unknowns and are solved during signal recovery.

The data is acquired using a radial ultrashort echo time (UTE) sequence with intermittent inversion pulses and a delay time for magnetization recovery. We acquire the data from 3T scanners, with and without motion.

\vspace{-1em}
\section{Proposed approach}
% \vspace{-1em}
\subsection{Signal Acquisition and Modeling}

\begin{figure}[ht]
	\centering	
	\includegraphics[width=0.49\textwidth,trim={0cm 0cm 18cm 0cm},clip]{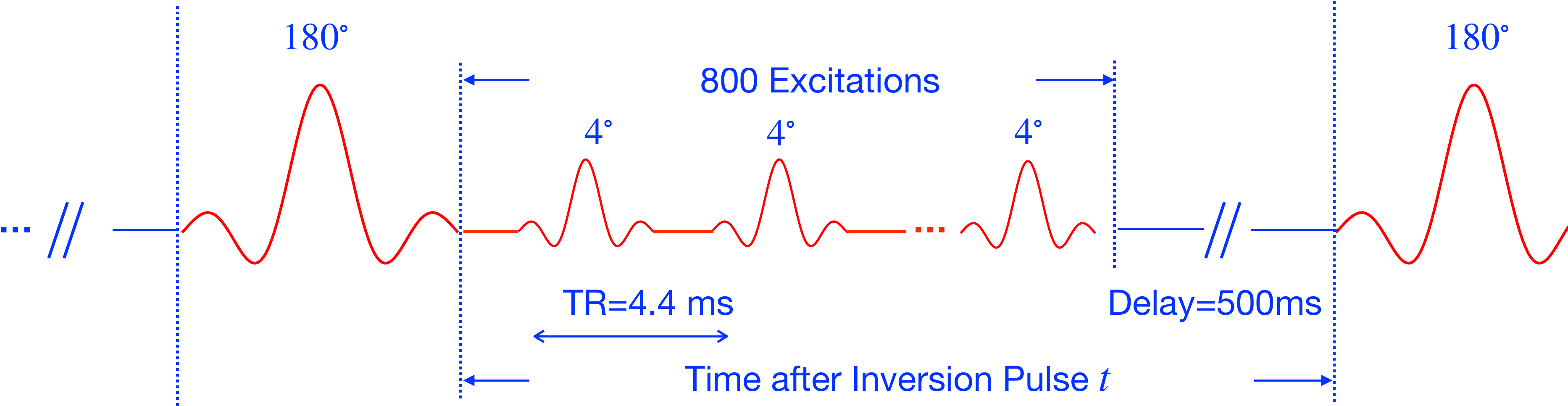}
	\caption{Illustration of the pulse sequence. The data is acquired using an inversion recovery with a segmented readout consisting of a train of UTE 3D radial acquisitions with $TR=4.4$ ms. Intermittent adiabatic inversion pulses are applied after 800 UTE radial lines, with a delay of 500 ms after each segmented acquisition block to allow for signal recovery to a steady state.  The total acquisition time was 4.3 minutes to acquire 50K spokes with a matrix size of 256$^3$. The radial spokes are ordered according to the tiny golden angle view order. The field of view was 24cm$^3$.}\vspace{-1em}
	\label{pulse}
\end{figure}

The acquisition of data is illustrated in Fig. \ref{pulse}. We assume that the evolution of magnetization during each inversion block is identical, and one initial inversion and segmented UTE scheme is performed to reduce steady-state effects. We consider the image volume $\boldsymbol{\rho}_{\tau}$ to be dependent on the delay $\tau(t)$ from the previous inversion pulse.

We model the signal measurements at the time instant
\begin{equation}
\label{static}
    \mathbf y_t = \mathbf A_t (\boldsymbol\rho_{\tau(t)}) + \mathbf n_t
\end{equation}
Here, $\mathbf A_t$ denotes the forward model that accounts for multichannel Fourier measurements and the k-space trajectory at time  $t$, which is measured from the start of the acquisition.  In the absence of motion, the measurements $y_t$ corresponding to a specific $\tau$ value can be pooled together. 

\vspace{-1em}
\subsection{Deep Factor Model}
\label{nomotion}
We note that low-rank factor models represent the signal as 
\begin{equation}
 \rho(\mathbf r,\tau) = \mathbf U(\mathbf r)~ \mathbf V(\tau),   
\end{equation}
where the columns of $\mathbf U(\mathbf r)$ are the spatial factors and the rows of $\mathbf V(\tau)$ are the temporal factors. The spatial basis functions are modulated by the temporal basis functions to generate the image at a specific time $\tau$. Note that low-rank factor models do not use any spatial regularization on the spatial basis functions $\mathbf U(\mathbf r)$.

We model the image $\boldsymbol\rho_{\tau}$ as the output of a deep conditional network $\mathcal N_{\boldsymbol \phi,\boldsymbol \theta}$
\begin{equation}
\label{dfm}
     \boldsymbol\rho(\mathbf r,\tau) = \mathcal N_{\boldsymbol \phi,\boldsymbol \theta}(\boldsymbol \gamma,\tau)
\end{equation}
$\boldsymbol \phi$ and $\boldsymbol \theta$ denote the parameters of $\mathcal N$, while $\tau$ denotes the inversion delay. $\boldsymbol \gamma$ is a coarse initial reconstruction. 
% We note that this $\boldsymbol \gamma$ may consist of images that are corrupted by extensive aliasing artifacts resulting from undersampling. 

We use a deep factor architecture to realize the conditional model as shown in Fig. \ref{fig:dfm}. In particular, we feed delay time $\tau$ to a dense network $\mathcal V_{\boldsymbol \theta}$, which provides the temporal factors
\begin{equation}
    \mathbf v(\tau) = \mathcal V_{\boldsymbol \theta}(\tau).
\end{equation}
We apply channel-wise multiplication between the feature maps and corresponding temporal factors to modulate the feature maps of $\mathcal U_{\boldsymbol \phi}$ and obtain 

% These factors are used to modulate the  channels of a convolutional neural network $\mathcal U_{\boldsymbol \phi}$ to obtain 
\begin{equation}
     \boldsymbol\rho(\mathbf r,\tau) = \mathcal U_{\boldsymbol \phi}\left(\boldsymbol \gamma(\mathbf r),\mathbf v(\tau)\right)
\end{equation}

One may view this approach as selectively activating and suppressing specific channels, based on the delay $\tau$. If the network $\mathcal U_{\boldsymbol \phi}$ only consists of a single hidden layer, this approach reduces to the traditional low-rank model. We expect the deep factorization at different layers to offer improved representation power. 
The unknowns of \eqref{dfm} are $\boldsymbol \phi$ and $\boldsymbol \theta$. In the absence of motion, we pose the reconstruction as 
\begin{equation}
	\{\boldsymbol \theta^*,\boldsymbol \phi^*\} = \arg \min_{\boldsymbol \theta,\boldsymbol \phi} \sum_t \|\mathbf y_t - \mathbf A_t\big(\mathcal N_{\boldsymbol \theta,\boldsymbol \phi}(\gamma,\tau)\big)\|^2
	\label{eq:dfm}
\end{equation}
% \textcolor{red}{The deep factor net architecture is an alternative for the generative model used in g-SToRM. The main distinction is the use of the input $\gamma(\mathbf r)$, which reduces the number of layers and trainable parameters, which is expected to translate to faster training. In addition, the factorization provides a more explainable architecture. } \textbf{It would be great if we compare our results with g-SToRM and low-rank models.}
\vspace{-1em}
\subsection{Deep Factor Model with Motion Compensation}
We assume the brain to be a rigid body and model any subject motion using 3 rotations and 3 translations. The object at time instant $t$ is assumed to be a rotated and translated version of $\boldsymbol\rho_{\tau(t)}$. We denote the transformation operator as $T_{\boldsymbol{\nu}_t,\boldsymbol \delta_t}$, where $\boldsymbol\nu_t$ and $\boldsymbol \delta_t$ are the rotation and translation parameters at time t. In particular, 
\begin{equation}
	T_{\boldsymbol{\nu},\mathbf \delta}\big(\rho(\mathbf r)\big) =  \boldsymbol\rho(R_{\boldsymbol\nu} \mathbf r + \boldsymbol\delta)
\end{equation}
where $\mathbf r$ is the coordinate in the image domain. With this assumption, \eqref{static} changes as
\begin{equation}
	\mathbf y_t = \mathbf A_t \big(T_{\boldsymbol{\nu_t},\boldsymbol \delta_t} \left(\boldsymbol\rho_{\tau(t)}\right)\big) + \mathbf n_t
\end{equation}
We note that a rotation in the image domain by an angle $\boldsymbol{\nu_t}$ corresponds to a rotation by the same angle in the Fourier domain. This amounts to rotating the k-space trajectory by the same angle. Similarly, a translation in the image domain corresponds to a phase modulation in the Fourier domain. Using these Fourier properties, we rewrite the above relation as 
\begin{equation}
	\mathbf y_t = \mathbf A_{t,\boldsymbol{\nu_t},\boldsymbol \delta_t} \big(\boldsymbol\rho_{\tau(t)}\big) + \mathbf n_t
	\label{eq:1}
\end{equation}
This reformulation allows us to absorb the impact of motion into the forward model and further extend the proposed method to DFM with motion compensation(DFM-MC). In the presence of motion, we use \eqref{eq:1} to minimize the cost function 
\begin{eqnarray}\nonumber
        C(\boldsymbol \theta,\boldsymbol \phi, \boldsymbol\nu, \boldsymbol\delta\} 
        &=&  \sum_t \|\mathbf y_t - \mathbf A_{t,\nu_t,\delta_t}\big(\mathcal N_{\boldsymbol \theta,\boldsymbol \phi}(\gamma,\tau)\big)\|^2 \\
        &&\qquad + \lambda_1\|\nabla {\boldsymbol \nu_t}\|^2+\lambda_2\|\nabla{\boldsymbol\delta_t\|^2}
        \label{eq:dfm with motion}
\end{eqnarray}
with respect to the network parameters $\boldsymbol \theta,\boldsymbol \phi$ as well as the motion parameters $\boldsymbol\nu, \boldsymbol\delta$.

\vspace{-1em}
\section{Implementation details}
% \vspace{-1em}

The data were acquired from two human volunteers on a 3.0T GE Premier scanner with 48 channel head coil. From each subject, we acquired two datasets. During the first acquisition, the subjects were instructed to stay still during the 4.3-minute acquisition. During the second acquisition, the subject was instructed to move the head multiple times during the scan. We implemented $\mathbf A$ using a multichannel NUFFT operator, which was differentiable with respect to the input image as well as the trajectory parameters.
% The original non-uniform Fast Fourier transform based on look-up tables was modified to make it differentiable. 
Post recovery, we fed the network $\mathcal N$ with different delays $\tau$ to generate the images. We used principal component analysis (PCA) to reduce the 48 coils to 10 virtual coils. 

\vspace{-1em}
\subsection{Deep Factor Model}

We implemented $\mathcal U_\phi$ using three 3D convolutional blocks, each with a 3D convolutional layer, a channel-wise multiplication layer, and a non-linear activation layer. Each block has 16 channels, except the last layer which has 2 channels representing the real and imaginary parts of the output image.  $\mathcal V_\theta$ contains two 2D convolutional blocks with kernel size 1 and 32 channels. The first block of $\mathcal U_\phi$ is activated by tanh function. Other blocks are activated by the Leaky ReLU function. $\phi$ and $\theta$ are randomly initialized.

We binned the data into eight subsets according to $\boldsymbol \tau(t)$ and gridding was applied to each subset to obtain 8 volumes, denoted by $\boldsymbol \gamma$. 
While these approximate reconstructions are very noisy and exhibit significant alias artifacts, they capture the contrasts reasonably well. 
In this work, we considered the recovery of images with 32 different inversion times; $\boldsymbol \tau(t)$ was linearly sampled from 0 to 1 with 32 steps. 
% $\boldsymbol \gamma$ and $\boldsymbol \tau(t)$ were not trainable.
We jointly optimized $\phi$ and $\theta$ based on Eq.\ref{eq:dfm}. In this section, we assume the acquisition was collected without motion. 

\vspace{-1em}
\subsection{Deep Factor Model with Motion Compensation}
We assume the motion after one inversion pulse remains to be the same and hence estimate the motion of the subject with temporal resolution=4s.  The parameters $\delta_t$ and $\nu_t$ were initialized as zeros in \eqref{eq:dfm with motion} and solved with $\theta$ and $\phi$.

\vspace{-1em}
\section{Results}
\subsection{Validation of Deep factor model}
\label{sec:validate dfm}

\begin{figure}[ht]
	\centering
	\includegraphics[width=0.47\textwidth]{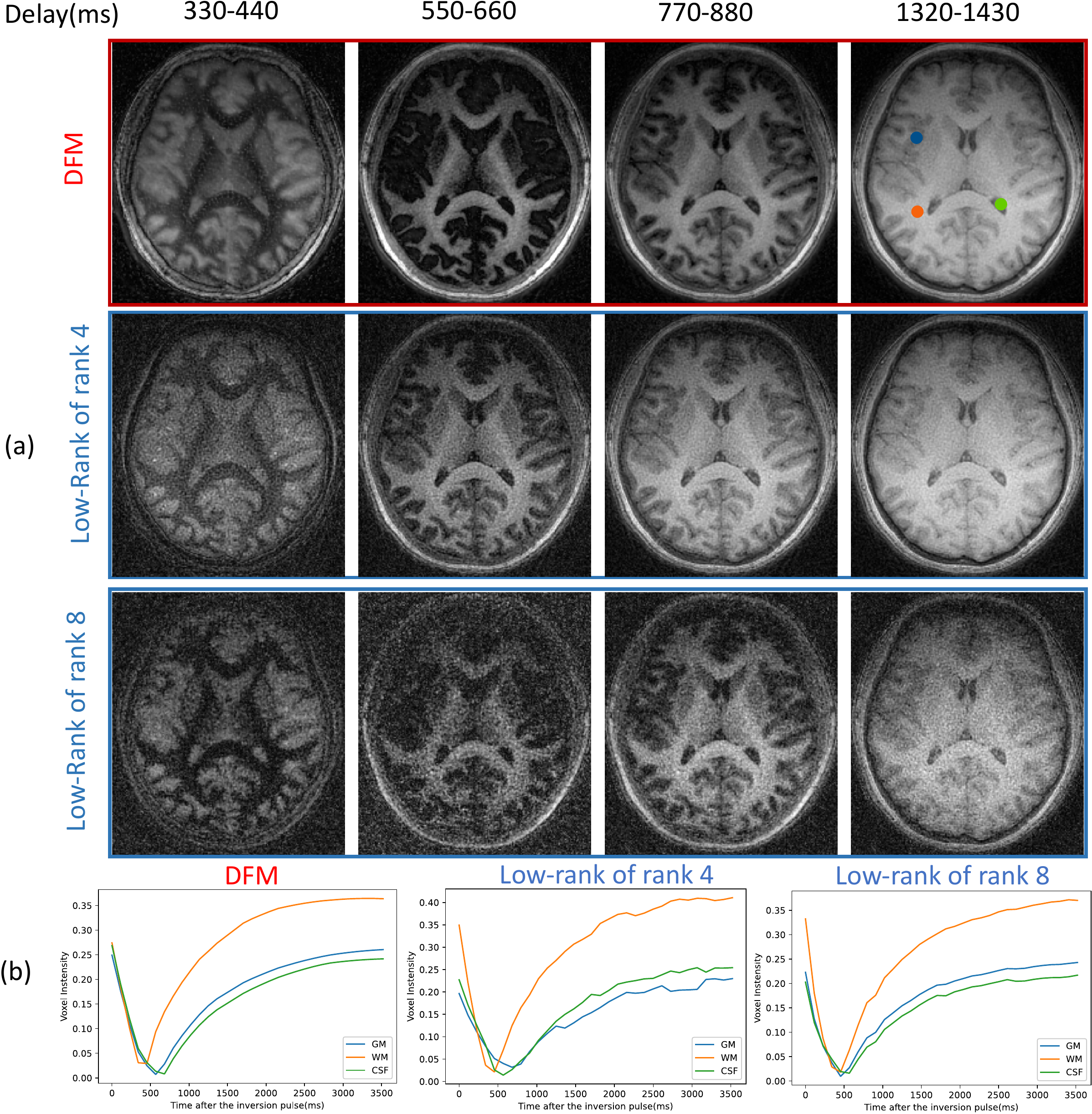}\vspace{-1em}
	\caption{Validation of DFM in the absence of motion. (a) The four columns correspond to four delay times $\tau$ after one inversion pulse; (b) Magnetization recovery curve of GM, WM, and CSF voxels indicated in the top image, estimated by different approaches.}
	\label{fig:dfm results}
\end{figure}
We first validate DFM in the absence of motion in Fig. \ref{fig:dfm results}. We compare DFM against the low-rank approach with two ranks. We observed that the lower rank translated to a poor fit to the magnetization recovery curve. In particular, we note that GM does not appear fully inverted in $r=4$ low-rank reconstructions at $\tau=550-660$ ms. By contrast, the DFM and $r=8$ low-rank reconstructions show nulled GM as expected. 

The reconstructed images demonstrate that the DFM can offer less noisy reconstructions, especially the bins corresponding to $\tau=330-440$ ms and $\tau=550-660$ ms, where WM and GM are nulled. We note that low-rank reconstructions exhibit significant noise, while the DFM reconstructions are relatively less noisy. The plots of the voxel intensity profiles, shown in Fig. \ref{fig:dfm results} (b),  show that the DFM curves are closer to the expected magnetization recovery. While the accuracy of the low-rank curves would improve with rank, this will come with increased sensitivity to noise.

\vspace{-1em}
\subsection{Validation of motion compensation}

\label{motion}
\begin{figure}[t!]
\centering
\includegraphics[width=0.47\textwidth]{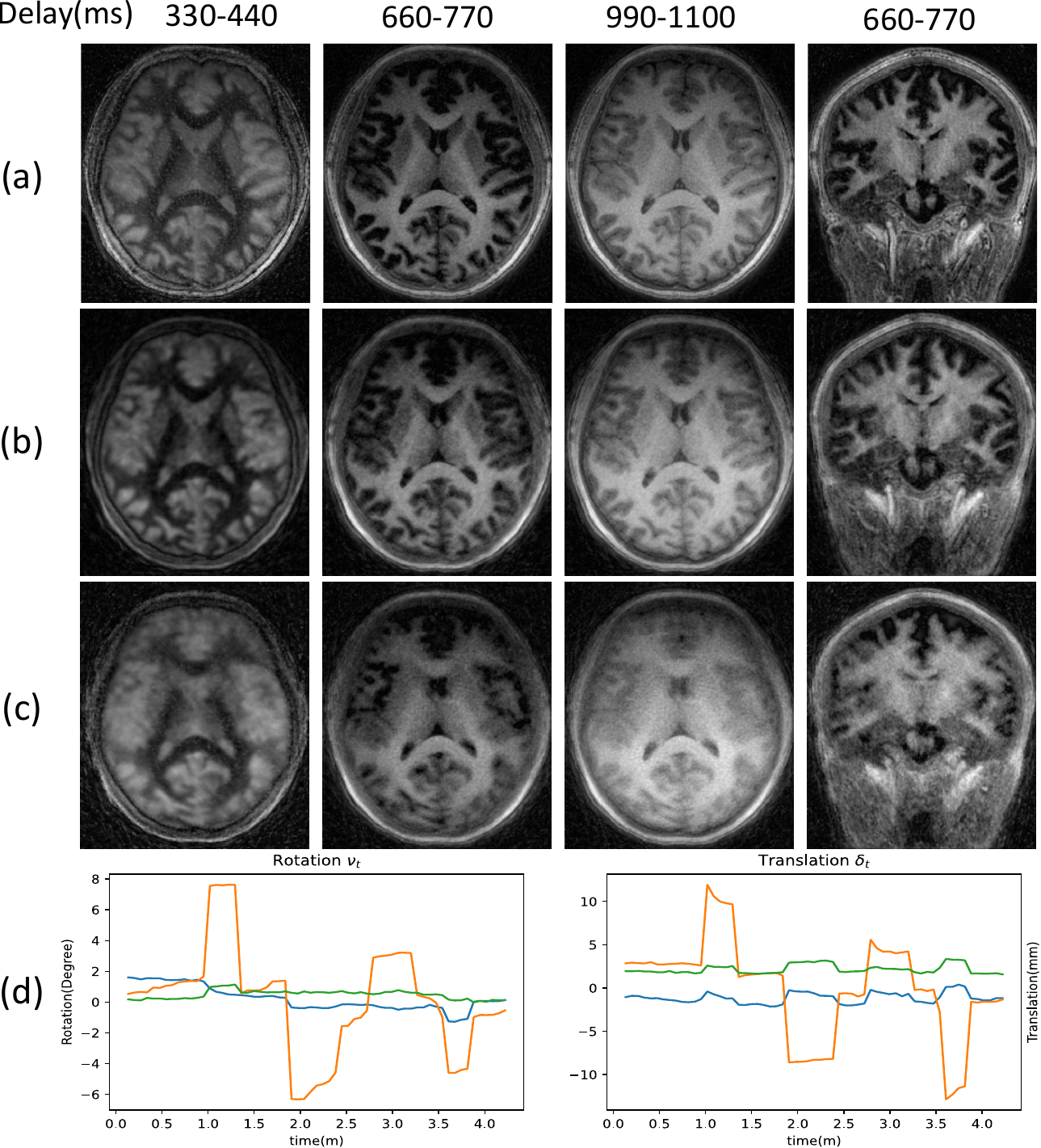}
\caption{Validation of DFM-MC. (a)(c) DFM reconstructions of the acquisitions with or without motion, respectively; (b) DFM-MC reconstructions of the acquisition with motion; (d) Motion parameters estimated by DFM-MC including rotation and translation to correct (b) from (c). }\vspace{-1em}
\label{fig:mc}
\end{figure}

The results in Fig. \ref{fig:mc} show the motion-compensated recovery using DFM-MC.
 The results demonstrate that the proposed DFM-MC can compensate for motion effects and offer reconstructions that are comparable to DFM reconstructions of the dataset without motion.

\vspace{-1em}
\section{Conclusion}
We introduced DFM to jointly recover multi-contrast images from inversion recovery MRI data. The proposed approach further extended and generalized the low-rank method and recovered a series of images. Our experiments demonstrate that the improved representation translated to higher-quality reconstructions than low-rank models. In particular, the DFM reconstructions are less noisy and offer a more faithful representation of the magnetization recovery. 
The reconstructions also show the great potential of modeling and correcting for motion during the acquisition, which would be beneficial while imaging older subjects. 
\vspace{-1em}
\section{Compliance with Ethical Standards}

This research study was conducted using human subject data. The institutional review board at the local institution approved the acquisition of the data, and written consent was obtained from all participants.

\bibliographystyle{IEEEbib}
\bibliography{refs}
\vspace{-1em}
\end{document}